\documentclass[letterpaper]{article} 
\usepackage{aaai25}  
\usepackage{times}  
\usepackage{helvet}  
\usepackage{courier}  
\usepackage[hyphens]{url}  
\usepackage{graphicx} 
\usepackage{natbib}  
\usepackage{caption} 
\usepackage{algorithm}
\usepackage{algorithmic}
\usepackage{subcaption}
\usepackage{dsfont}
\usepackage{array}
\usepackage{colortbl, xcolor}
\usepackage{newfloat}
\pdfoutput=1
\usepackage{listings}
\DeclareCaptionStyle{ruled}{labelfont=normalfont,labelsep=colon,strut=off} 
\floatstyle{ruled}
\newfloat{listing}{tb}{lst}{}
\floatname{listing}{Listing}
\title{Measuring and Mitigating Hallucinations in Vision-Language Dataset Generation \\ for Remote Sensing}
\author{
    Madeline Anderson\textsuperscript{\rm 1},
    Miriam Cha\textsuperscript{\rm 2},
    William T. Freeman\textsuperscript{\rm 1},
    J. Taylor Perron\textsuperscript{\rm 1},
    Nathaniel Maidel\textsuperscript{\rm 3},\\
    Kerri Cahoy\textsuperscript{\rm 1}
}
\affiliations{
    \textsuperscript{\rm 1}MIT,
    \textsuperscript{\rm 2}MIT Lincoln Laboratory,
    \textsuperscript{\rm 3}DAF MIT AI Accelerator

}

\usepackage{bibentry}

\begin{document}

\maketitle

\begin{abstract}
Vision language models have achieved impressive results across various fields. However, adoption in remote sensing remains limited, largely due to the scarcity of paired image-text data. To bridge this gap, synthetic caption generation has gained interest, traditionally relying on rule-based methods that use metadata or bounding boxes. While these approaches provide some description, they often lack the depth needed to capture complex wide-area scenes. Large language models (LLMs) offer a promising alternative for generating more descriptive captions, yet they can produce generic outputs and are prone to hallucination. In this paper, we propose a new method to enhance vision-language datasets for remote sensing by integrating maps as external data sources, enabling the generation of detailed, context-rich captions. Additionally, we present methods to measure and mitigate hallucinations in LLM-generated text. We introduce fMoW-mm, a multimodal dataset incorporating satellite imagery, maps, metadata, and text annotations. We demonstrate its effectiveness for automatic target recognition in few-shot settings, achieving superior performance compared to other vision-language remote sensing datasets.
\end{abstract}

%

\section{Introduction}

In recent years, there have been significant advancements in vision-language models, leading to powerful applications across many fields \cite{vlsurvey_zhang, vlsurvey_long, vlsurvey_du}. However, adoption within the remote sensing community has lagged, largely due to the limited availability of paired data for remote sensing imagery and text. Recently, researchers have started to address this gap by generating synthetic captions for remote sensing images \cite{diffusionsat, remoteclip, GeoRSCLIP}. Traditionally, rule-based methods leveraging metadata \cite{diffusionsat} and bounding boxes \cite{remoteclip} have been used, but these approaches fall short when it comes to fully describing the complexity of wide-area remote sensing scenes.

The adoption of large language models (LLMs) offers a promising alternative, as LLMs can potentially generate more descriptive and contextually rich captions \cite{GeoRSCLIP}. Yet, LLM-generated text for remote sensing data often remains generic, and importantly, is prone to hallucination. This issue of hallucination has yet to be thoroughly explored in the context of vision-language dataset for remote sensing, where accurate and detailed scene descriptions are important for data curation.

\begin{figure}[t]
    \centering
    \includegraphics[width=\linewidth]{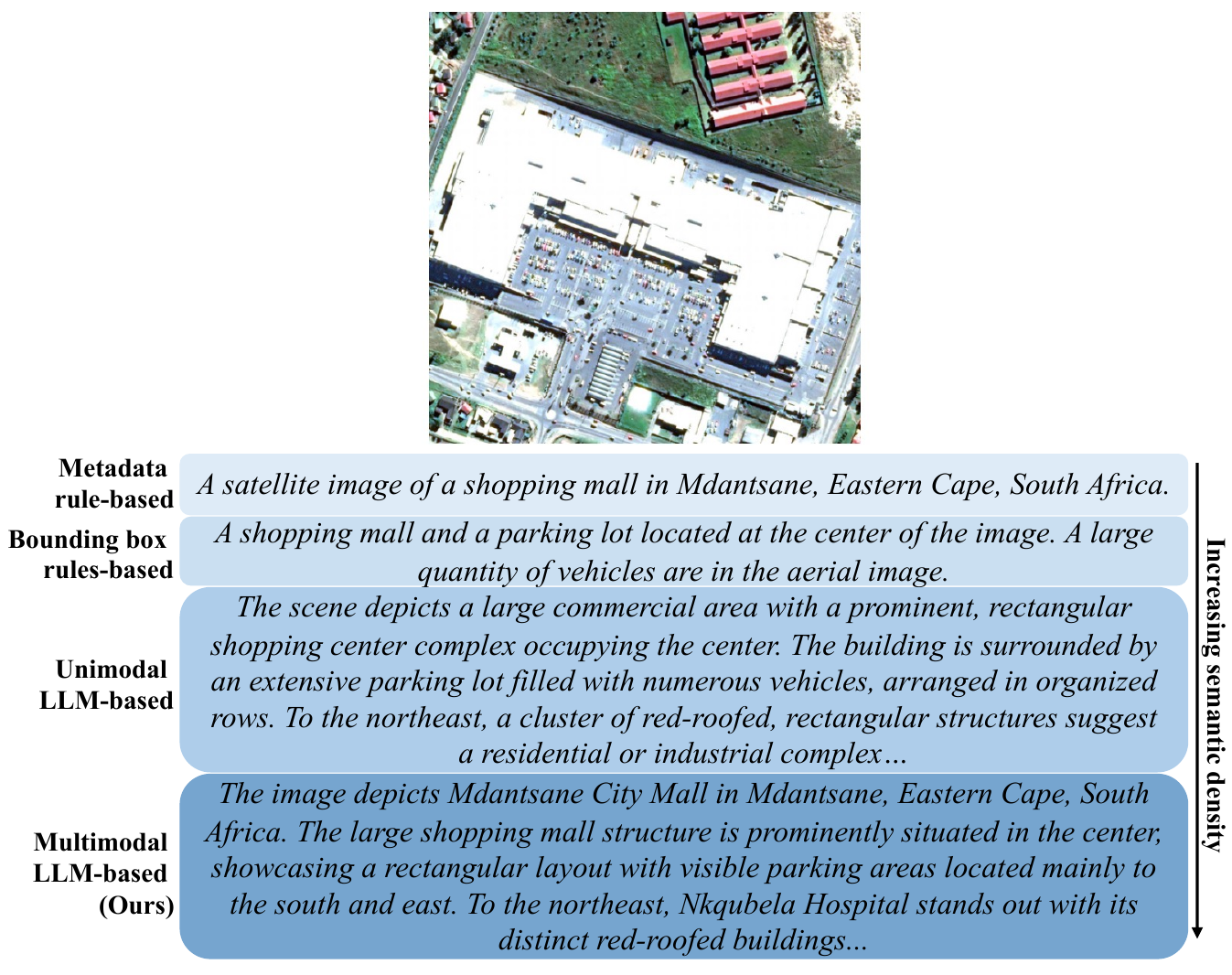}
    \caption{\textbf{Comparison of captioning methods:} Rule-based captions are limited in detail. Unimodal LLM captions are fluid but often generic. Wide-area scenes covering diverse structures and objects require semantically rich descriptions. We leverage the semantic density of maps to generate comprehensive and detailed captions.}
    \label{fig:motiv}
\end{figure}

\renewcommand{\arraystretch}{1.5}
\begin{table*}[t]
\centering
\scriptsize
\begin{tabular}{|>{\centering\arraybackslash}m{3.6cm}|>{\centering\arraybackslash}m{2.9cm}|>{\centering\arraybackslash}m{5.5cm}|>{\centering\arraybackslash}m{3cm}|}
\hline
\rowcolor[gray]{0.9}
\textbf{Dataset Name} & \textbf{Captioning Method} & \textbf{Caption Quality} &\textbf{External Info} \\ \hline
RSVQA \cite{lobry2020rsvqa} & Map data rules-based & Focuses on reasoning & Maps (OSM) \\ \hline
Skyscript \cite{wang2023skyscript} & Map data rules-based & Rigid and limited to OSM tags & Maps (OSM) \\ \hline
DiffusionSAT \cite{diffusionsat} & Metadata rules-based & Rigid with details limited to metadata & Metadata \\ \hline
RemoteCLIP \cite{remoteclip} & Bounding box rules-based &  Rigid with details limited to bounding box objects & None \\ \hline
ChatEarthNet \cite{yuan2024chatearthnet} & LLM-based & Fluid sounding with details limited to landcover & Landcover (WorldCover) \\ \hline
RS5M \cite{GeoRSCLIP} & Web-filtered, LLM-based & Coarse detail with focus on central objects & None \\ \hline
\textbf{fMoW-mm (Ours)} & \textbf{Multimodal LLM-based} & \textbf{Fluid sounding with comprehensive, specific details} & \textbf{Metadata + maps (OSM)} \\ \hline
\end{tabular}
\caption{Overview of various vision-language remote sensing datasets.}
\label{table:vldatasets}
\end{table*}

In this paper, we propose a new approach for curating vision-language datasets for remote sensing by integrating external sources of information, such as maps and metadata. Maps offer a rich source of contextual information, including labels and segmentation maps. Using these external sources, we introduce a method to generate more comprehensive and detailed captions for remote sensing images than existing methods allow (Figure~\ref{fig:motiv}).

To address the issue of hallucinations, we present methods to measure and mitigate hallucination in LLM-generated captions. Additionally, we introduce \textbf{fMoW-mm}, a new multimodal dataset (built upon fMoW \cite{christie2018functionalmapworld}), which includes satellite imagery, maps, metadata, and text annotations. Finally, we demonstrate the effectiveness of fMoW-mm in automatic target recognition under few-shot conditions, showcasing the potential of this dataset for enhancing remote sensing applications with limited labeled data. Our contributions are as follows:
\begin{itemize}
  \item We introduce a novel dataset curation method that leverages external data sources, specifically maps, for enhanced language descriptions of remote sensing images.
  \item We present fMoW-mm, a comprehensive multimodal dataset cross-referenced with fMoW, consisting of satellite imagery, map, metadata, and text annotations.
  \item We explore methods to measure and mitigate hallucinations in LLM-generated captions for remote sensing.
  \item We demonstrate the utility of fMoW-mm for automatic target recognition in limited-label scenarios.
\end{itemize}

\section{Related Work}

\subsection{Vision-Language Datasets for Remote Sensing}
Although large vision-language datasets are less common in the remote sensing domain, several have been developed in recent years. We review six existing datasets in Table \ref{table:vldatasets}. Datasets that rely on rules-based captions, such as RSVQA \cite{lobry2020rsvqa}, Skyscript \cite{wang2023skyscript}, DiffusionSAT \cite{diffusionsat}, and RemoteCLIP \cite{remoteclip}, often produce rigid captions with limited detail. The content is constrained by the external information fed into the rules-based frameworks, such as OpenStreetMap (OSM) data for RSVQA and Skyscript, metadata for DiffusionSAT, and bounding boxes for RemoteCLIP. ChatEarthNet \cite{yuan2024chatearthnet} and RS5M \cite{GeoRSCLIP} leverage LLMs, resulting in more fluid sounding captions. However, ChatEarthNet captions primarily describe landcover, while RS5M captions, generated using BLIP-2 \cite{li2023blip2bootstrapping}, contain coarse details. RS5M also includes internet-scraped image-text data, often centered on a single object, which may not represent typical remote sensing images. We address these shortcomings by leveraging a multimodal LLM (GPT-4o) with multiple sources of external information (maps and metadata) to generate comprehensive, detailed, and fluid captions for complex remote sensing scenes.

\subsection{Hallucination Metrics and Mitigation Strategies}

Measuring and mitigating hallucinations in LLM-generated captions is critical. Existing hallucination metrics include statistical, model-based, and vision-language measures \cite{ji2023survey}. Statistical metrics like ROUGE \cite{rouge}, BLEU \cite{papineni2002bleu}, and PARENT \cite{dhingra2019handling} assess hallucinations based on n-gram overlaps. Model-based metrics include Information Extraction \cite{singh2018natural}, QA-based methods \cite{deutsch2021question}, Natural Language Inference \cite{dusekkasner2020evaluating}, and Faithfulness Classification \cite{liu2022token}, and LM-based approaches \cite{filippova2020controlled}. However, many rely on task-specific datasets or LLM access, which may not be available. Metrics specific to vision-language hallucinations are very scarce \cite{rohrbach2019object}. To address the lack of suitable metrics, we propose a statistical metric inspired by BLEU precision that uses OSM tags as source text to measure hallucination rates.


LLM hallucination mitigation strategies include data-based, modeling-based, and post-processing methods \cite{ji2023survey}. Data-based strategies include caption ranking or filtering and information augmentation with synthetic or external data. Modeling techniques include planning and sketching \cite{wang2021sketch}, reinforcement learning \cite{uc-cetina2022survey}, multi-task learning \cite{weng2020towards}, and controllable generation \cite{rashkin2021increasing, wu2021controllable}. Post-processing focuses on correcting hallucinations after captions are generated. Without direct LLM access required by many modeling methods, we mitigate hallucinations using data-based strategies (external data augmentation) and post-processing techniques (prompt ensembling).
\section{Multimodal Dataset Curation}

\begin{figure}
    \centering
    \includegraphics[width=\linewidth]{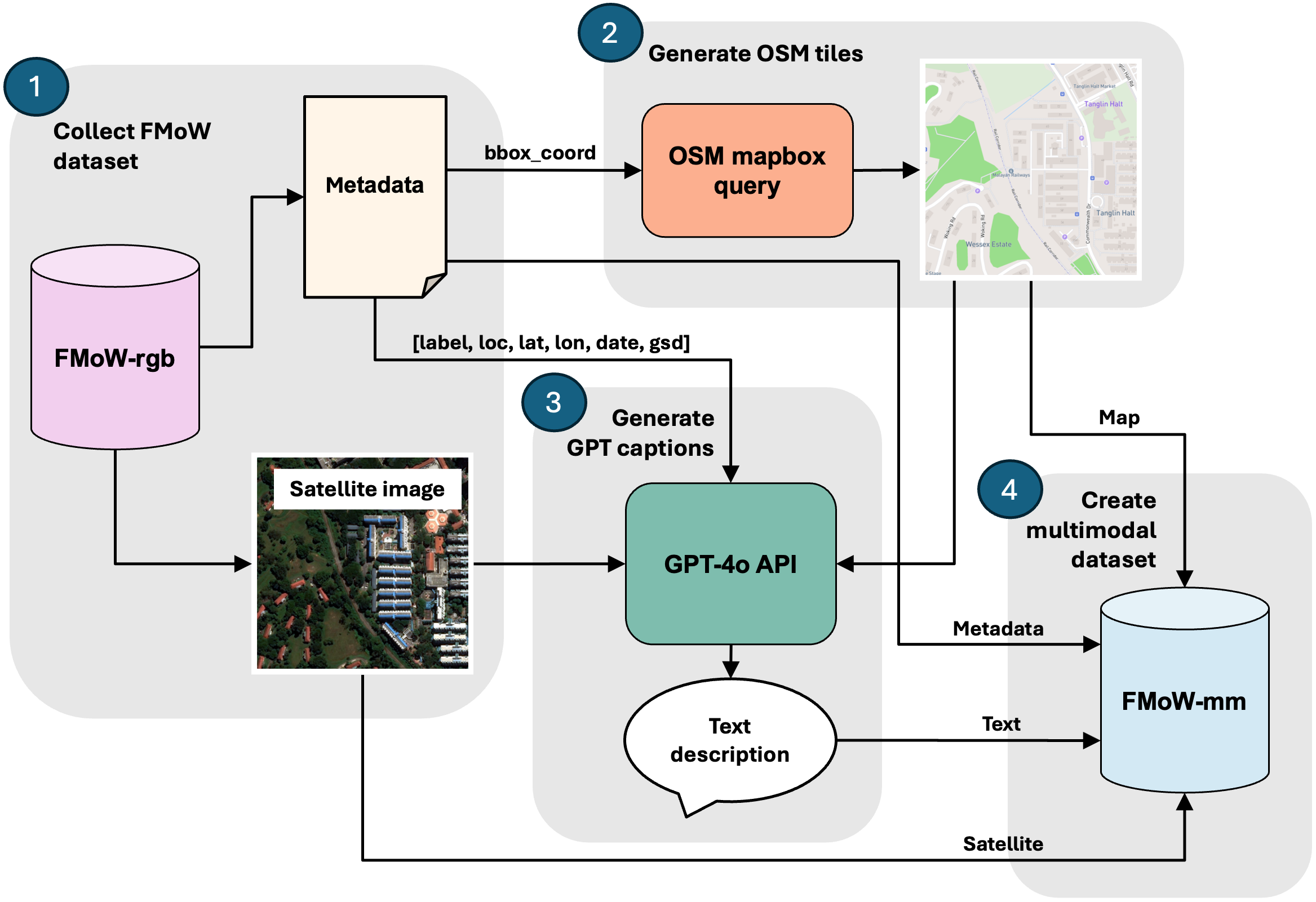}
    \caption{fMoW-mm data curation pipeline}
    \label{fig:pipeline}
\end{figure}

\begin{figure*}
    \centering
    \includegraphics[width=0.95\linewidth]{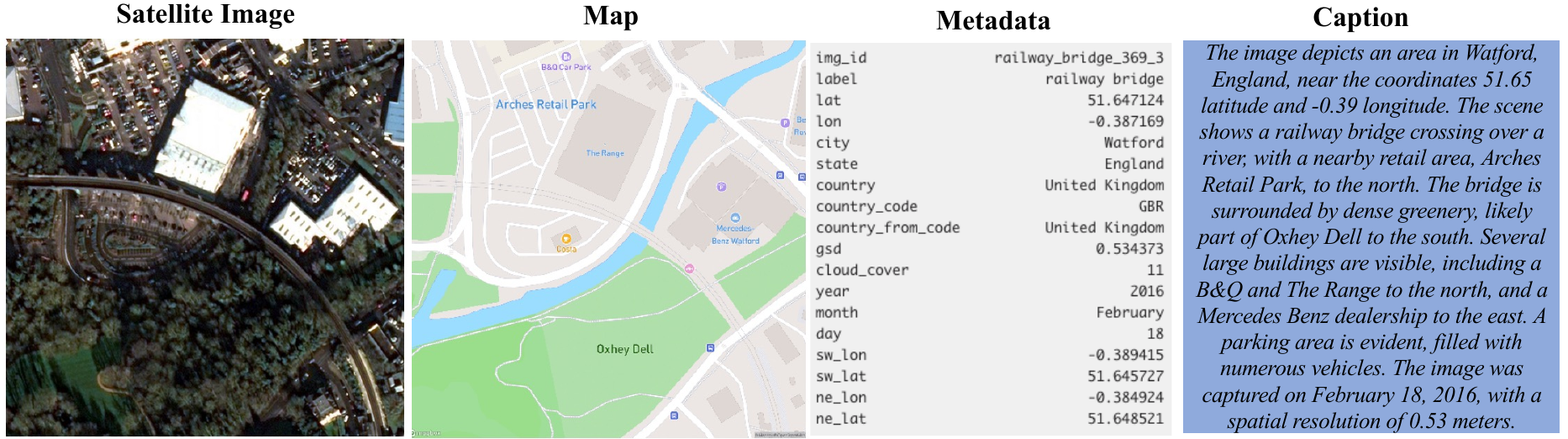}
    \captionsetup{width=0.95\linewidth}
    \caption{A sample from the fMoW-mm dataset. The generated caption accurately incorporates information from the satellite image, map, and metadata.}
    \label{fig:fmow_mm}
\end{figure*}

Figure~\ref{fig:pipeline} outlines the fMoW-mm  curation process: 1) Gather satellite images and metadata from fMoW-rgb, 2) Use bounding box metadata to perform an OSM Mapbox query and retrieve map tiles, 3) Input satellite images, maps, and metadata into GPT-4o to generate captions, 4) Combine these elements to create fMoW-mm. Each step is detailed in the following subsections.

\subsection{Functional Map of the World (fMoW-rgb)}
The fMoW-rgb dataset consists of 83,412 remote sensing images that feature objects in 63 categories \cite{christie2018functionalmapworld}. Each image comes with corresponding metadata such as category label, latitude, longitude, timestamp, ground sampling distance (GSD), and bounding box. 


    
\subsection{OpenStreetMap (OSM) Tile Retrieval}
We use the bounding box coordinates from the fMoW-rgb metadata to query the corresponding OSM Static Image tiles through the Mapbox API. Map styles are customized using the online Mapbox studio.
    
\subsection{Caption Generation with GPT-4o}
To generate captions, we use the GPT-4o API from OpenAI, which accepts visual and text inputs. For each sample, we input the fMoW-rgb satellite image, metadata and OSM tile. The input metadata includes the category label, location (city, state/region, country), latitude, longitude, and GSD. We prompt GPT-4o to describe the remote sensing scene and to include landmarks, relative positions, sizes, colors, and quantities, while leveraging the metadata and map for context. 
Other LLMs, including open-source options, can be substituted for GPT-4o, as long as they accept visual inputs. 


\subsection{Multimodal Functional Map of the World (fMoW-mm)}


We combine the fMoW-rgb satellite image and metadata, the OSM tile, and the GPT-4o generated caption to create 83,412 tuples of \{satellite, metadata, map, text\}. Figure~\ref{fig:fmow_mm} shows a sample from the fMoW-mm dataset. The full dataset is available at \url{https://bit.ly/fMoW-mm}. 

\section{Hallucination Metric}



Hallucinations often occur when the LLM infers incorrect landmarks during caption generation. To quantify these hallucinations, we compute the false discovery rate (FDR), inspired by BLEU precision, which measures the proportion of false positives in the generated text. Unlike BLEU, which evaluates n-gram overlaps, we calculate precision over variable-length proper nouns and define FDR as $1-precision$:
\begin{equation}
\small
FDR = 1 - \frac{\sum_{c \in C}\mathds{1}_{R}(c)}{K}
\end{equation}
where the candidate list $C = [c_1, c_2, ..., c_{K}]$ is an array of $K$ proper nouns, and the reference list $R = [r_1, r_2, ..., r_{M}]$ is an array of $M$ proper nouns. The indicator function $\mathds{1}_{R}(c) = 1$ if $c\in R$ and 0 otherwise. FDR reflects the proportion of false positives among all predicted positives, quantifying the rate of hallucinations in the generated (candidate) captions.



\begin{figure}[t]
    \centering
    \includegraphics[width=\linewidth]{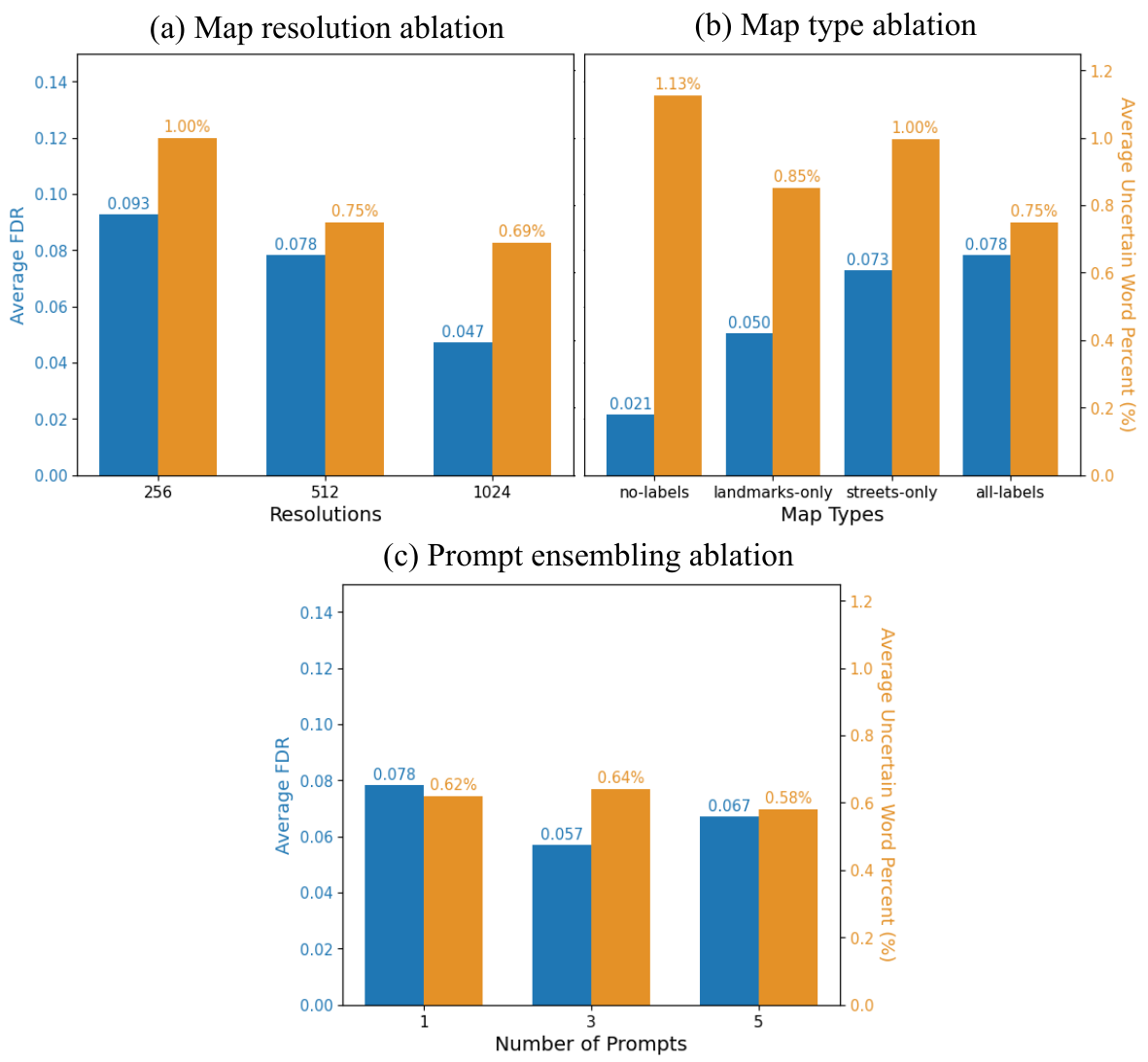}
    \caption{\textbf{Ablations. (a) Map Resolution:} Higher resolution reduces hallucination rates and uncertainty in generated captions. \textbf{(b) Map Types:} Using landmarks-only gives the best balance, reducing hallucinations while limiting uncertainty. \textbf{(c) Prompt Ensembling:} Combining captions from multiple prompts did not significantly impact the metrics, however increasing from 3 to 5 prompts may result in repeated hallucinations that propagate into the final caption.}
    \label{fig:ablation_exp}
\end{figure}


\section{Experiments}
We perform ablations to evaluate how components of our curation pipeline affect hallucination rates (FDR) and measure the percentage of uncertain words as a proxy for LLM uncertainty. We then demonstrate fMoW-mm's effectiveness in enhancing few-shot object detection performance.

\subsection{Ablations}
\begin{itemize}
    \item \textbf{Map Resolution:} We vary the resolution of the OSM input to GPT-4o, considering \{256, 512, 1024\}.
    \item \textbf{Map Types:} We explore four map variations:
    \begin{itemize}
        \item All Labels: Includes all available labels on the map.
        \item Landmarks-Only: Includes only landmark labels, excluding street names.
        \item Streets-Only: Includes only street names, excluding landmark labels.
        \item No Labels: Displays the segmentation map without any text labels.
    \end{itemize}
    \item \textbf{Prompt Ensembling:} We generate multiple prompts for the same question and aggregate the responses to analyze convergence. We experiment with \{1, 3, 5\} prompts.
\end{itemize}

Figure~\ref{fig:ablation_exp} shows that increasing map resolution reduces hallucination rates and uncertain word percentages, highlighting the importance of map legibility. We use a $1024\times1024$ resolution for fMoW-mm. While further increases in resolution may offer additional benefits, we leave this exploration for future work due to computational constraints.

Adding text labels, such as landmarks and street names, predictably increases hallucination rates. The inclusion of street names (e.g., streets-only, all-labels) results in a more pronounced increase, likely because non-horizontally aligned street names introduce ambiguity that leads to hallucinations. Landmark names, which are consistently horizontal, cause fewer issues. Captions generated without labels (i.e., no-label) achieve the lowest hallucination rates but are often overly generic, with a high rate of uncertain word usage. For the fMoW-mm dataset, we selected the landmarks-only configuration as it strikes a good balance, minimizing hallucinations while maintaining reasonable specificity.

Prompt ensembling did not result in noticeable improvements. We suspect that repeated hallucinations across responses may increase overlap, propagating errors into the final captions. For fMoW-mm, we aggregate responses from three prompts, yielding the lowest FDR.










\subsection{Few-Shot Object Detection with CLIP}

We continually pretrain the CLIP \cite{radford2021learning} ViT-L/14 model using the fMoW-mm dataset and evaluate the learned visual representation on few-shot object detection \cite{bou2024exploring}. The model was continually trained for 50 epochs with a batch size of 125. We compare performance with vision-language baselines: CLIP, OpenCLIP, GeoRSCLIP, and RemoteCLIP. 

Table \ref{table:ovdsatresults} shows the mAP50 scores for 5, 10 and 30-shot detection on the DIOR dataset \cite{li2020object}, averaged over 5 splits. Our model demonstrates improved performance across all n-shots, showing its viability for data-scarce scenarios. Although the fMoW-mm dataset is much smaller than the datasets used for GeoRSCLIP (RS5M, \raisebox{0.5ex}{\texttildelow}5M) and RemoteCLIP (\raisebox{0.5ex}{\texttildelow}150k), it achieves superior performance, highlighting the benefits of increased semantic density in the generated captions. To isolate the impact of the dataset, comparisons are limited to CLIP models.

\begin{table}[t]
\footnotesize 
\setlength{\tabcolsep}{4pt} 
\centering
\begin{tabular}{cccc}
\hline
\textbf{Backbone} & \textbf{5-shot} & \textbf{10-shot} & \textbf{30-shot} \\ \hline \hline
CLIP \cite{radford2021learning} & 0.1447 & 0.1872 & 0.1810 \\
OpenCLIP \cite{cherti2023reproducing} & 0.1477 & 0.1863 & 0.1804 \\
GeoRSCLIP \cite{GeoRSCLIP} & 0.1401 & 0.1791 & 0.1815 \\
RemoteCLIP \cite{remoteclip} & 0.1571 & 0.1893 & 0.1903 \\
\textbf{Ours} & \textbf{0.1574} & \textbf{0.1902} & \textbf{0.1972} \\ 
\hline
\end{tabular}
\caption{mAP50 scores for 5, 10, and 30-shot object detection on the DIOR dataset using various visual backbones with ViT-L/14, averaged across 5 splits.}
\label{table:ovdsatresults}
\end{table}


\section{Conclusion}
In this work, we explored methods to measure and mitigate hallucinations in captions describing remote sensing imagery. Previous approaches to caption generation have often resulted in rigid and generic descriptions. Our approach enhances vision-language datasets in remote sensing by integrating maps as external data sources, enabling the creation of more detailed and contextually rich captions. Through the introduction of fMoW-mm—a multimodal dataset extending the fMoW dataset with satellite imagery, maps, metadata, and text annotations—we demonstrate a reduced rate of hallucinations and improved performance in automatic target recognition under few-shot conditions.
\section{Acknowledgments}

{\fontsize{9.5}{1}\selectfont 
Research was sponsored by the Department of the Air Force Artificial Intelligence Accelerator and was accomplished under Cooperative Agreement Number FA8750-19-2-1000. The views and conclusions contained in this document are those of the authors and should not be interpreted as representing the official policies, either expressed or implied, of the Department of the Air Force or the U.S. Government. The U.S. Government is authorized to reproduce and distribute reprints for Government purposes notwithstanding any copyright notation herein.
}

\bibliography{aaai25.bib}

\end{document}